\pdfoutput=1
\documentclass[11pt]{article}

\usepackage[final]{acl}
\usepackage{amssymb}
\usepackage{amsmath}
\usepackage[ruled]{algorithm2e}
\usepackage{amsfonts}

\usepackage{times}
\usepackage{latexsym}
\usepackage{graphicx}
\usepackage[T1]{fontenc}
\usepackage[utf8]{inputenc}

\usepackage{microtype}

\usepackage{inconsolata}
\usepackage{algpseudocode}
\usepackage{multirow}
\usepackage{tikz}
\usetikzlibrary{positioning}
\title{Pose2Gest: A Few-Shot Model-Free Approach Applied In South Indian Classical Dance Gesture Recognition}

\author{Kavitha Raju \\ IIIT-Kottayam \\  Bridge Connectivity Solutions  \And Nandini J. Warrier \\  IIIT-Kottayam\\  \And Manu Madhavan \\ IIIT-Kottayam \AND Selvi C.\\ IIIT-Kottayam \And Arun B. Warrier\\ Kerala Kalamandalam \And  Thulasi Kumar\\ Kerala Kalamandalam }

\begin{document}
\maketitle
\begin{abstract}
 The classical dances from India utilize a set of hand gestures known as Mudras, serving as the foundational elements of its posture vocabulary. Identifying these mudras represents a primary task in digitizing the dance performances. With \textit{Kathakali}, a dance-drama, as the focus, this work addresses mudra recognition by framing it as a 24-class classification problem and proposes a novel vector-similarity-based approach leveraging pose estimation techniques. This method obviates the need for extensive training or fine-tuning, thus mitigating the issue of limited data availability common in similar AI applications. Achieving an accuracy rate of 92\%, our approach demonstrates comparable or superior performance to existing model-training-based methodologies in this domain. Notably, it remains effective even with small datasets comprising just 1 or 5 samples, albeit with a slightly diminished performance. Furthermore, our system supports processing images, videos, and real-time streams, accommodating both hand-cropped and full-body images. As part of this research, we have curated and released a publicly accessible  Hasta Mudra dataset, which applies to multiple South Indian art forms including \textit{Kathakali}. The implementation of the proposed method is also made available as a web application.

\end{abstract}

%\keywords{Gesture recognition, Indian traditional dance, Pose Estimation, Normalization, Vector Similarity Search}

\section{Introduction} 

\begin{figure*}
    \centering
    \includegraphics[width=\textwidth]{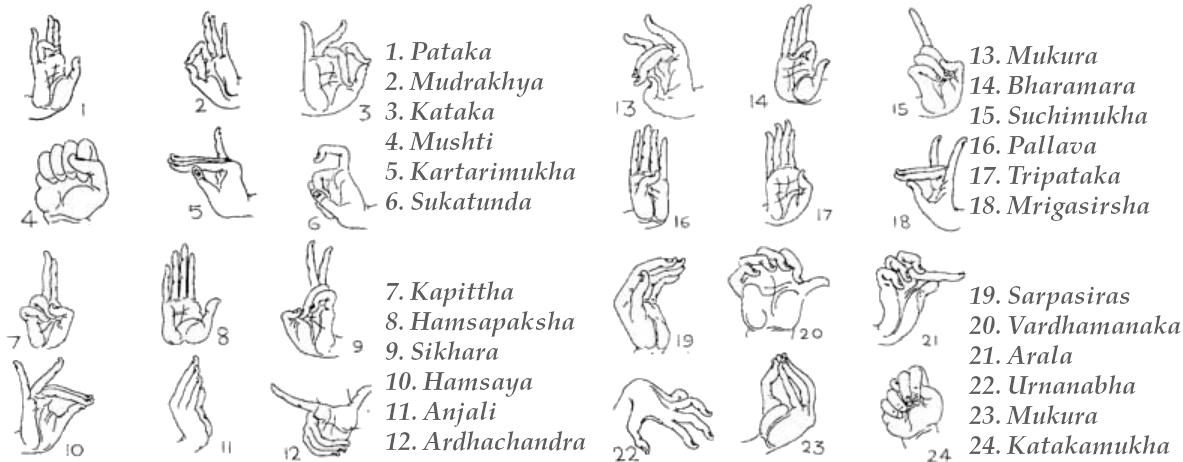}
    \caption{The 24 mudra classes in Hasta Mudra: Common for the dance forms \textit{Kathakali, Kutiyattam, Mohiniyattam, Krishnanattam,} etc.\cite{langKathakali}}
    \label{fig:mudras}
\end{figure*}

India boasts a diverse array of dance and theatrical art forms. The classical dances of India are highly revered forms of artistic expression that have deep cultural and historical significance. \textit{Kathakali}, along with several other traditional performing art forms like \textit{Kutiyattam, Mohiniyattam, Krishnattam}, etc. rely heavily on a specialized set of hand gestures known as \textbf{Mudras} to communicate with the audience. Mudras form the cornerstone of their intricate dance movements and poses. Mudras common to these art forms are depicted in Fig \ref{fig:mudras}. 

The aforementioned theatre/dance forms namely \textit{Kutiyattam, Kathakali, Mohiniyattam, Krishnattam} are a few classical art forms of India predominantly performed in the southern state of Kerala.  \textit{Kutiyattam} is one such ancient theatre system that was recognized as an intangible cultural heritage by UNESCO in 2008\cite{unesco_kutiyattam}. The performers of these art forms communicate with the audience through facial expressions (\textit{Abhinaya}), hand gestures (\textit{Mudra}), and dance poses. The performer on the stage is generally accompanied by music and percussion instruments to accentuate the performance. While sharing common foundational mudras, these art forms diverge significantly in the mudra utilization for constructing the more complex semantic elements of the dance system, such as meaningful words and sentences. Also, there are other art forms like \textit{Bharatanatyam, Kuchipudi,} etc. from other regions of Southern India which have the same concept of mudra, but the names, shapes, and the number of mudras differ from that of \textit{Kathakali}.

In this paper, we have focussed on \textit{Kathakali}- one of the traditional classical art forms mentioned prior as the primary focus of the discussion but the application of the proposed methodology is extendable to other art forms like \textit{Kutiyattam, Mohiniyattam, and Krishnattam}. 
\textit{Kathakali}, which originated somewhere in the 17th century is a dance drama, where a story is acted out using an exquisite and well-defined set of hand gestures, facial expressions along with rhythmic poses and body movement. The hand gestures (mudras) used in \textit{Kathakali} have evolved out of other ancient art forms that prevailed even before \textit{Kathakali} and the authoritative treatises on the Indian dance like \textit{Hastalakshanadeepika, Natyasatra, Abhinaya darpana,} etc. The mudras form the language of \textit{Kathakali}\citep{langKathakali}. On one hand, the mudras form a language system that has the capability of expressing even the minute details of the story in a well-defined and structured manner, whereas, on the other hand, when attempting to computationally model this mudra language, we can observe the common challenges of natural language processing like ambiguity, diverse dialects, etc.  Even outside of the computational aspects, the language of \textit{Kathakali} is so intricate and elaborate that it requires expert knowledge for a person to understand the performance.

Accurately identifying the mudras represents a crucial initial step in digitally processing these dance/theatre performances. There are a basic 24 mudras, which form the alphabet of this mudra language. Further, words and sentences of this language are developed using the basic 24 mudras. In this study, as a first step towards digitally processing \textit{Kathakali}, we try to classify the basic mudras into 24 distinct mudra classes.

The basic 24 mudras when used in specific manners(body posture, combination, and movements), can be interpreted as meaningful words of the language\citep{hastalakshana}. mudras are formed by holding the fingers of a hand in a specific pose. These hand poses can be held in different angles and orientations or moved in different directions and still be identified as the same mudra. The concept of such mudras is commonly seen in other Indian classical dance forms and they share a lot of similarities. But the mudras system of \textit{Kathakali} is more structured and systematic, almost like an independent unique sign language.  

Processing stage-performed art forms digitally poses numerous challenges due to the complex nature of the raw data, often captured as images and videos. Converting this data into a format amenable to automatic processing and comprehension involves overcoming several obstacles. These include identifying the performer, tracking their finger, face, and body movements, recognizing gestures and spoken words, and deciphering the meaning conveyed by the combination and sequence of these elements. Ultimately, the goal is to translate this narrative into written languages such as English. Addressing these challenges requires a sophisticated technological framework, encompassing computer vision techniques like gesture recognition and natural language processing methods such as language understanding and translation. Deep learning plays a crucial role in enabling these functionalities, leveraging extensive training data and computational infrastructure. However, sourcing such comprehensive training datasets for specific use cases like stage-performed art forms can be daunting.

Moreover, even if a dataset is constructed for one particular art form, adapting the same approach to another form restarts the process from scratch, hindering efficiency and scalability. Given the imperative of preserving endangered knowledge systems and cultural heritage inherent in these art forms, ease of adaptation becomes paramount. Consequently, the ability to operate with minimal data becomes essential, facilitating broader applicability and ensuring the preservation of diverse cultural expressions.

To break down the complex task of automatically interpreting \textit{Kathakali} performances, a logical starting point is to focus on recognizing its fundamental elements, such as the mudras, which serve as the basic units of the art form. In this study, we propose a method for identifying mudras in \textit{Kathakali} performances, leveraging a limited number of training samples. This approach offers the potential for broad applicability, as the same techniques could be easily adapted to other dance forms or related domains, such as sign language recognition, by constructing a reference dictionary from a small set of example images.

This methodology not only represents an initial step towards developing an interpretation system for \textit{Kathakali} but also holds promise for various other applications. For instance, it can facilitate the creation of additional training data by analyzing unannotated video footage of performances. Furthermore, it could serve as a valuable educational tool, supporting digital training initiatives in the realm of \textit{Kathakali} and similar art forms.

Our innovative approach, based on vector-similarity and pose estimation techniques, eliminates the need for extensive model training or fine-tuning, addressing the common issue of limited data availability in similar AI applications. With an impressive accuracy of 92\%, our method demonstrates performance comparable to or surpassing existing model-based approaches in this field. Moreover, our technique remains effective even when dealing with small datasets comprising just 1 or 5 samples, albeit with slightly reduced accuracy. Additionally, our system is versatile, capable of processing various forms of input, including images, videos, and real-time streams, and can handle both hand-cropped and full-body images. 

Pose2Gest makes use of existing general-purpose pose estimation technologies and uses that information to build vector representations of the gesture/mudra classes we have. These vectors are formed using Euclidean coordinates of different landmarks of the hand, after applying necessary normalizations. Now the test data is also subjected to similar vectorization and then compared against our database of known mudra vectors to find the one it is most similar to. The pose estimation model used is Mediapipe from Google which can estimate 3D coordinates of hand landmarks from images, videos, and even real-time feeds\citep{lugaresi2019mediapipe}.

The key contributions of the work are the following:
\begin{enumerate}
    \item Pose2Gest: A novel approach is proposed that can perform image and video classification for dance gesture recognition.
    \item Evaluation of the proposed method by comparing it with existing works as well as in data-constrained settings.
    \item Hasta Mudra Dataset: A dataset is prepared and made available publically at \url{https://github.com/kavitharaju/KathakaliMudraDataset} with all the 24 mudras used in dances like \textit{Kutiyattam, Kathakali, Mohiniyattam, Krishanattam,} etc.
    \item A web-based software,\url{https://mudraresearch.iiitkottayam.ac.in/},  that implements the Pose2Gest approach and serves as a means for crowd-sourcing data collection.
\end{enumerate}
The remaining parts of this article include a review of existing literature, the proposed methodology, the experiments conducted comparing the proposed Pose2Gest approach with available datasets of prior works in the domain, the details of the Hasta Mudra dataset made available as part of this work, the experiments conducted on the Hasta Mudra dataset, the web application that implements the proposed Pose2Gest approach,  discussions and conclusion.

\section{Related Works}

\textit{Kathakali}, and the other traditional art forms in general, are not areas of wide popularity in this digital era, and hence technological research is scarce. That is indicative of the relevance of such research explorations and the existing gap. Among the existing works in \textit{Kathakali} and other related fields, one common trend seen is that all such attempts rely on building deep learning models from the ground up by building datasets first which is the major challenge of this domain. Our work on the other hand aims to use minimum data, leverage the powers of existing models, and apply it to multiple tasks. Such an approach is novel in the explored literature.

In a work done by Bhavanam et al. for \textit{Kathakali} Hand Gesture Recognition  (\citet{bhavanam2020classification}), which attempted the same task and achieved an accuracy of 74\% by applying Convolutional Neural Networks on a dataset of 654 images cropped down to show only one hand. The dataset (\citet{neelakanta_iyer-dataset}) developed and published along with the work, comprises color images of 56x56 pixels. Authors have compared the performance of the SVM classifier on the same dataset and found CNN to be outperforming it. We tested the Pose2Gest methodology on the dataset developed by Bhavanam et al.(\citet{neelakanta_iyer-dataset})  and observations are given in the results section.

In a separate \textit{Kathakali} Mudra Recognition work (\citet{malavath2023classification}) upon the same dataset (\citet{neelakanta_iyer-dataset}) as above, an improved accuracy of 79\% was reported, which was again implemented using CNN models. The work also provided a comparison with a Naive Bayes model.

As part of a framework to computationally analyze \textit{Kathakali} Videos (\citet{bulani2022framework}), a pose estimation model that can detect body landmarks on a \textit{Kathakali} performance video is built. This was essential as the general purpose pose estimation systems will not work on \textit{Kathakali} images because of its vibrant costumes and face painting. Further, the capabilities of such a pose estimation system for style transfer and avatar puppeteering of the \textit{Kathakali} model, synthesizing cartoon videos of \textit{Kathakali} performances, etc. were explored. This work showcased the effectiveness of pose-estimation-based processing and its scope. In Pose2Gest, the pose estimation module used is easily replaceable by other similar technologies like these when adapting to input in costumes and facial expression recognition.

The \textit{Kathakali} character(\textit{Vesham}) recognition work (\citet{iyer2024kathakali}) dealt with a related problem in the same area. It was concerned with identifying the role of the performer based on his attire. This was also treated as an image classification task using deep learning techniques such as CNN and reports a top accuracy of 97.5\% through rigorous parameter tuning and experiments. The work on \textit{Kathakali} face expression detection (\citet{selvi2021kathakali}), also deals with a different problem of classification based on facial expression in the same domain of \textit{Kathakali} Analysis.

Mudras used by different art forms could be similar but need not be exactly the same. \textit{Bharatanatyam} is another such classical dance form in India that has wider popularity compared to \textit{Kathakali}. It also has the concept of mudras, though the hand shapes are not exactly the same as that used in \textit{Kathakali} and other dance/ theatre forms from Kerala. In a recent work (\citet{parthasarathy2023novel}), a benchmark dataset of 5 mudra classes was published and a method for classification of mudras using MobileNetV2 models after fine-tuning was proposed. The work had real-time and offline performance tests. In the offline tests where they used 6000 images for 5 classes, the reported accuracy of 86.45\%. Pose2Gest has also been tested against this dataset to evaluate its adaptability to a problem other than \textit{Kathakali}.

Another \textit{Bharatanatyam} dance related work (\citet{vadakkot2023automatic}), reported an accuracy of 92\% upon using CNN for \textit{Bharatanatyam} mudra recognition considering 29 classes of mudras. Further, an approach of using Eigen Mudra projections was also proposed. This work built a dataset of 68,073 high-resolution color images, from 6 subjects. The higher accuracy of this work compared to the \textit{Kathakali} works (\citet{bhavanam2020classification, malavath2023classification}) could be attributed to the bigger dataset and better quality images, which might have also required higher computational resources. Finetuning of pre-trained models like VGG19 reported an improved performance of 94\%, which showed the effectiveness of building on top of general-purpose models.

At a similar task of Indian Classical dance mudra recognition, (\citet{kumar2018indian}) a different approach of using a Histogram and other such features and employing an SVM classifier were explored. It was also a 24-class classification task for the dance form \textit{Kuchipudi}. This work utilized the pre-deep-learning approaches of more mathematically aware feature extraction and showed the effectiveness of working with interpretable features, a smaller size dataset of 25 samples per class, and a simpler ML algorithm. Their reported mudra recognition frequency was 89\%.

Other works in Indian classical dance mudra recognition, (\citet{pradeep2023recognition, haridas2022bharathanatyam}) also employ CNN. The latter used YOLO for \textit{Bharatanatyam} mudras and reported an accuracy of 73\%. The former evaluated multiple CNN models. This work also tried to locate hands in bigger images and then classify them, which caused a lower accuracy of 50\%. Other works (\citet{nandeppanavar2023bharathanatyam}) that attempted using VGG-19 and ResNet50V2 and reported 96.44\% accuracy, also exhibited a similar pattern of depending on deep learning architectures for either finetuning or training from scratch.
 
The dance pose recognition works on K-Pop (\citet{kim2017classification}) utilized pose detection as its feature. But for pose detection, it used Kinect which required a studio environment to capture that data. But with the present-day advancements in pose estimation models which can work on images and videos,  this limitation is eliminated. The work used a dataset of 800 data points for 200 classes, which is as low as 4 samples per class, and indicated the effectiveness of the pose-estimation-based approach on a smaller dataset. Multiple machine learning algorithms were tried on these features, like SVM, KNN, ELM, etc.

A slightly more diverse application domain of pose recognition for human-robot interaction (\citet{gao2021robot}), adapted the use of hand pose estimation. Previously hand-glove-based approaches have shown good performance in such tasks. Now those principles are applied using pose estimates from images/videos via openpose \cite{cao2019openpose}. They used a 3D-CNN and an LSTM to further process these pose features and reported the highest accuracy of 92.4\%.

Apart from relying on large training datasets, a common issue with many CNN-based approaches discussed previously is their use of cropped hand images. When applied to classifying mudras in video data, adapting these methods requires substantial additional effort to locate hands within larger frames and process continuous video data efficiently.

\begin{figure*}
    \centering
    \includegraphics[width=0.95\textwidth]{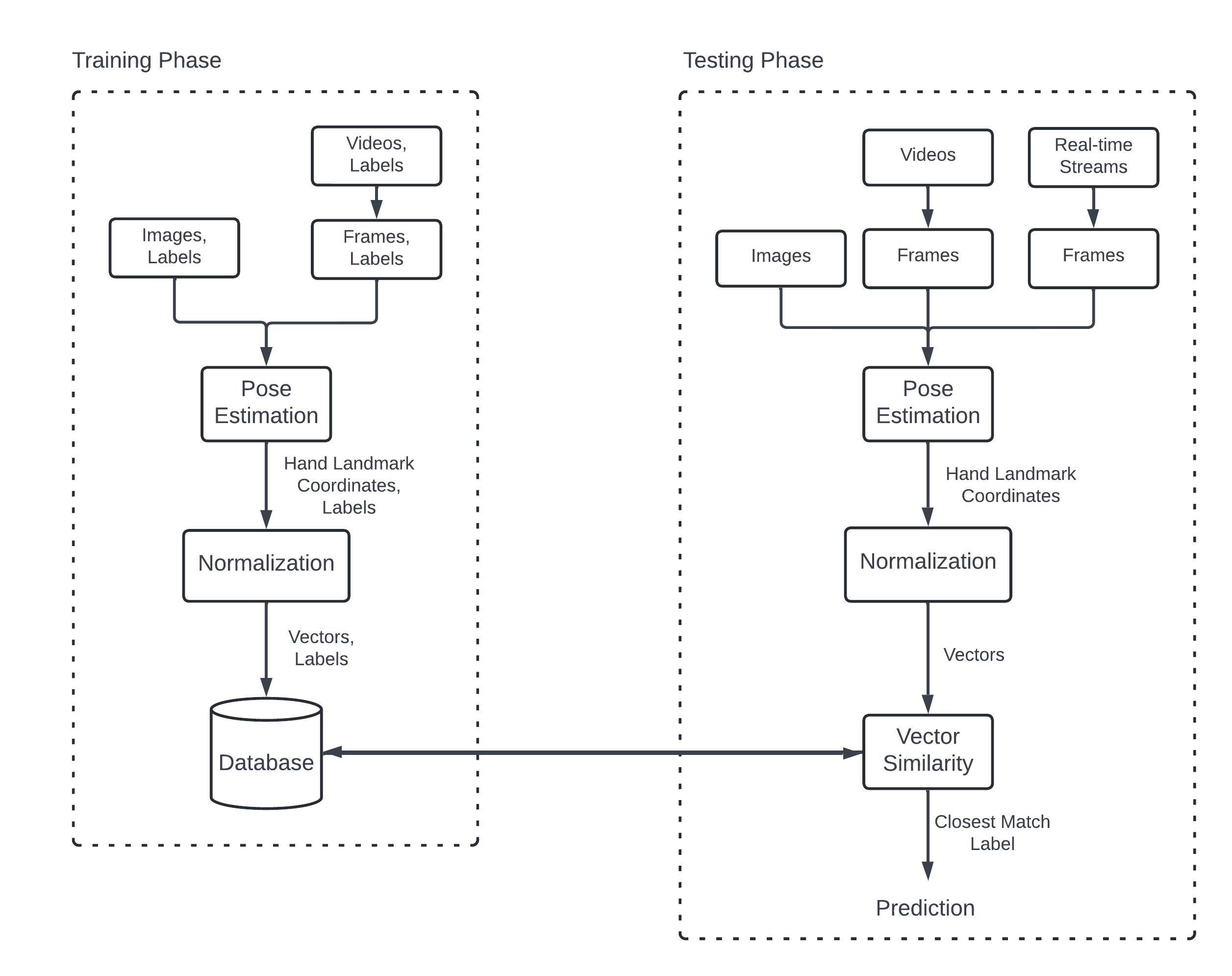}
    \caption{Processes involved in mudra Recognition using Pose2Gest}
    \label{fig:block}
\end{figure*}

\section{Methodology}

The Pose2Gest approach expects the raw data as images or frames of videos. The images are subjected to feature extraction using a pose estimation system and the obtained features are normalized for performing vector similarity for class prediction. These processes are depicted in Fig \ref{fig:block} and each is elaborated in the following sub-sections.

\subsection{Data}
The data required for the proposed work require good quality images or videos with the hands showing the mudra being clearly visible. These media files are expected to belong to one of the 24 classes of the \textit{Kathakali} mudras, shown in Fig \ref{fig:mudras}. Video inputs are transformed into frames for processing, and the method is efficient enough to ensure speedy execution. The approach can work in a low-resource setting as low as 1 sample image per class for training. This would lower the accuracy, but this performance can still be helpful in cases where a small drop in prediction accuracy is acceptable. For instance, there's a probability that some of the frames in a video sequence analysis will have inaccurate mudra results. One technique to deal with this is to consider a window size and use the most predicted class as the output to mitigate the fluctuations. Another choice is to use a language model to predict the most likely mudra sequence based on each frame's top N classes. Under such circumstances, a small loss in accuracy can be tolerated.

In the training set comprising five images per class, we enhanced the variety of hand orientations. This orientation difference is illustrated in Fig \ref{fig:norm}. The dataset now encompasses images featuring left and right hands, close-up and distant shots, as well as various angles of left- and right-facing palms. This broader coverage in the training set enhances accuracy, particularly for mudras where certain finger positions may not be visible from specific angles.

Another notable feature of the system is its versatility in handling input data. Despite being trained on samples featuring only isolated hand portions, the system can effectively process videos or images depicting the full upper body or extensive background scenery. As long as the input data maintains sufficient clarity for accurate pose detection of hand landmarks, the system's training and testing configurations remain largely independent of each other.

\subsection{Pose Estimation}
Pose estimation is a computer vision technique where a neural network model is trained to detect important points or landmarks of an object. For instance, given the image of a box, identify the corners and provide the x, and y coordinate values for them in the image. In our use case, we rely on a pose estimation system for detecting hand landmarks given the image of a person.

There are multiple pre-trained models with good accuracy and efficiency trained for general-purpose human pose estimation. Openpose \citep{cao2019openpose} and Google's Mediapipe \citep{lugaresi2019mediapipe} are two prominent pose estimation frameworks. In this work, we have utilized Mediapipe, due to factors like its cross-platform support, lightweightedness, and efficiency, even though it is easily replaceable by a different model like Openpose or one trained specifically for \textit{Kathakali} 
 \citep{bulani2022framework}. 

\begin{figure*}
    \centering
    \includegraphics[scale=0.35]{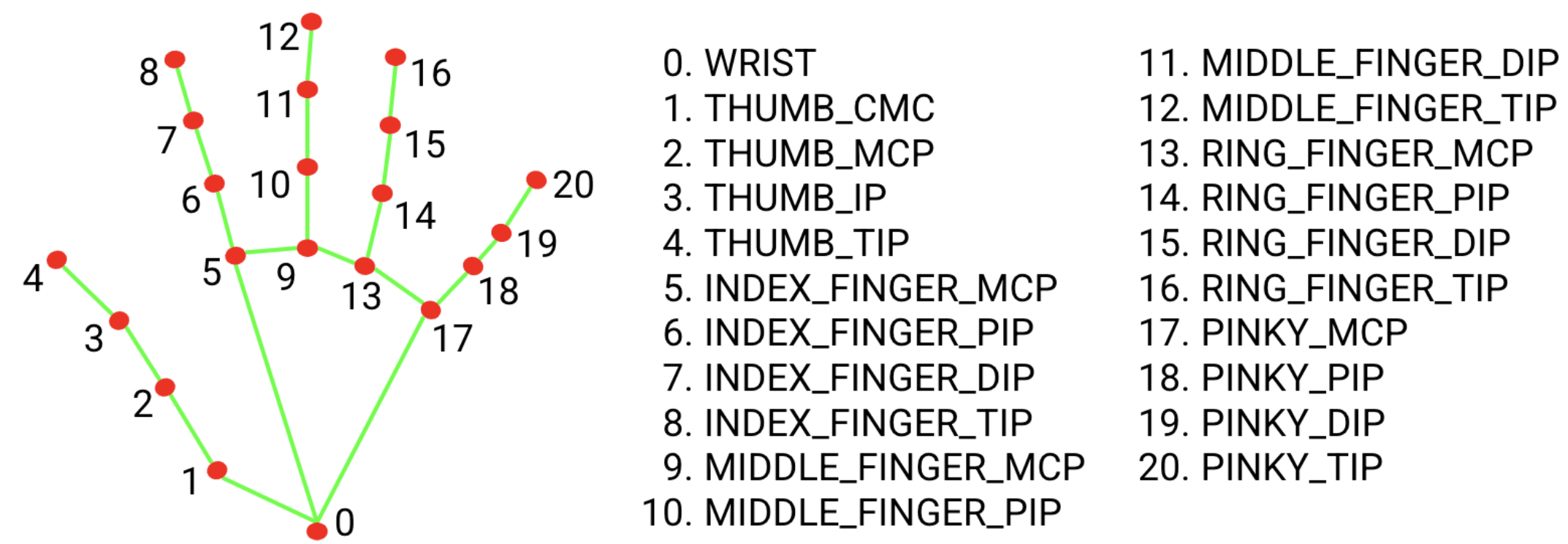}
    \caption{Hand landmarks identified by Mediapipe\citep{mediapipe_hands}}
    \label{fig:hand}
\end{figure*}

\subsection{Features} 

The common approach of using image pixel values as features for training CNNs from scratch or fine-tuning pre-trained image classification neural networks would all require large amounts of labeled training data to automatically deduce the relevant features and converge with reasonable accuracy. But even then, those models will be highly sensitive to changes in training and test data. Another option commonly adopted that makes use of existing pre-trained models is to use the embeddings from those models as our features. These lack the advantage of being interpretable. 

The features used in the proposed system are the coordinate value outputs obtained from pre-trained pose estimation models. Here, there is no training or fine-tuning involved, rather we use the existing general-purpose model as such. It gives the added advantage of being human-interpretable, which allows them to be subjected to post-processing tasks like normalization as per our requirements.

The raw images obtained from video frames, real-time feeds, or image datasets are passed on to the pose estimation model and we obtain the hand landmarks output of that model. In this implementation, we have relied on the hands model\citep{mediapipe_hands} Mediapipe provides. This enables us to get a jump start concerning the computer vision aspects of the task. Also helps in eliminating a lot of irrelevant features like backgrounds, person-dependent features, etc, and focuses only on what is relevant for us in our task, giving a more robust system deployable in diverse settings than what it has been trained on. Mediapipe gives  3D coordinates values of 21 landmarks of hands thus giving us a 63 dimension feature vector for an image. Fig \ref{fig:hand} shows the 21 points on one hand that Mediapipe can detect. For each of these points $(x, y, z)$ coordinate values, as per image frame will be obtained, giving rise to a $21\times3=63$ dimension vector.

\subsection{Normalization}

Normalization refers to the process of standardizing data values within a certain range. It is commonly used to bring data into a consistent format that aids in comparison, analysis, etc. As we include hands with different orientations, the 
$x$, $y$, and $z$ coordinates obtained for hand in the image frame, need not be directly comparable with each other. To bring the hand coordinates to a uniform size, location, and angle, we subject it to normalization. This becomes even more critical as we aim to give consistent performance but provide only very little data for it to generalize over the diversities.

The normalization applied in our methodology is the projection of the coordinate obtained from the pose estimation system, which is based on the input image frame, to a fixed coordinate system such that this transformation may involve rotation, scaling, and translation of the points. For a 3-dimensional point, this can be achieved by multiplying it with a 4x4 transformation matrix \citep{linearalgebra}. For a landmark in the image coordinate system $s_i = (s_{i,x}, s_{i,y}, s_{i,z})$, we need a transformation matrix $T \in \mathbb{R}^{4\times 4}$, such that:
\[p_i = s_i\cdot T\]
where $p_i = (p_{i,x}, p_{i,y}, p_{i,z})$ is the corresponding point for the landmark in the normalized coordinate system. The locations of the reference landmarks in the normalized coordinate system are fixed, whereas the source coordinates would vary from image to image. Therefore the transformation, $T$, need not be the same for all the images. Hence we propose a dynamic method for computing the transformation matrix based on a fixed set of reference points in the target coordinate system.

\begin{figure}
    \includegraphics[width=0.45\textwidth]{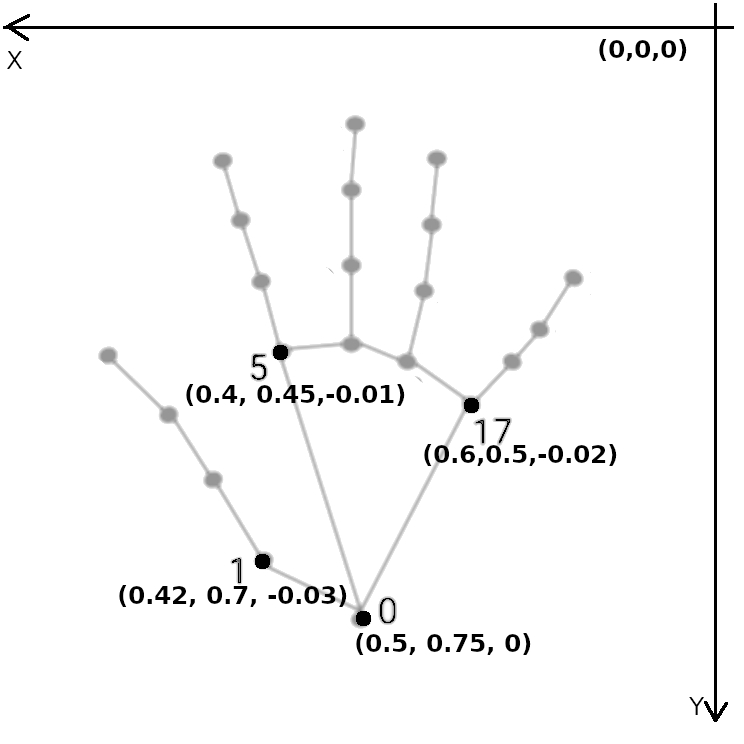}
    \caption{The four reference landmarks on the hand used for comparing the fixed values in the primary system and input values in the secondary system, to determine the required information are highlighted.}
    \label{fig:P-values}
\end{figure}

The technique used to derive this transformation matrix involves utilizing four key reference points, shown in Fig \ref{fig:P-values}: the wrist, the base points of the index finger, the pinky finger, and the thumb. These points are selected because they have the least relative movement between each other. There will be two sets of values for these points: the fixed primary set that we initialize with values as per our need and the secondary set which is obtained from pose estimation on input images.

\begin{figure}
    \centering
    \includegraphics[width=0.4\textwidth]{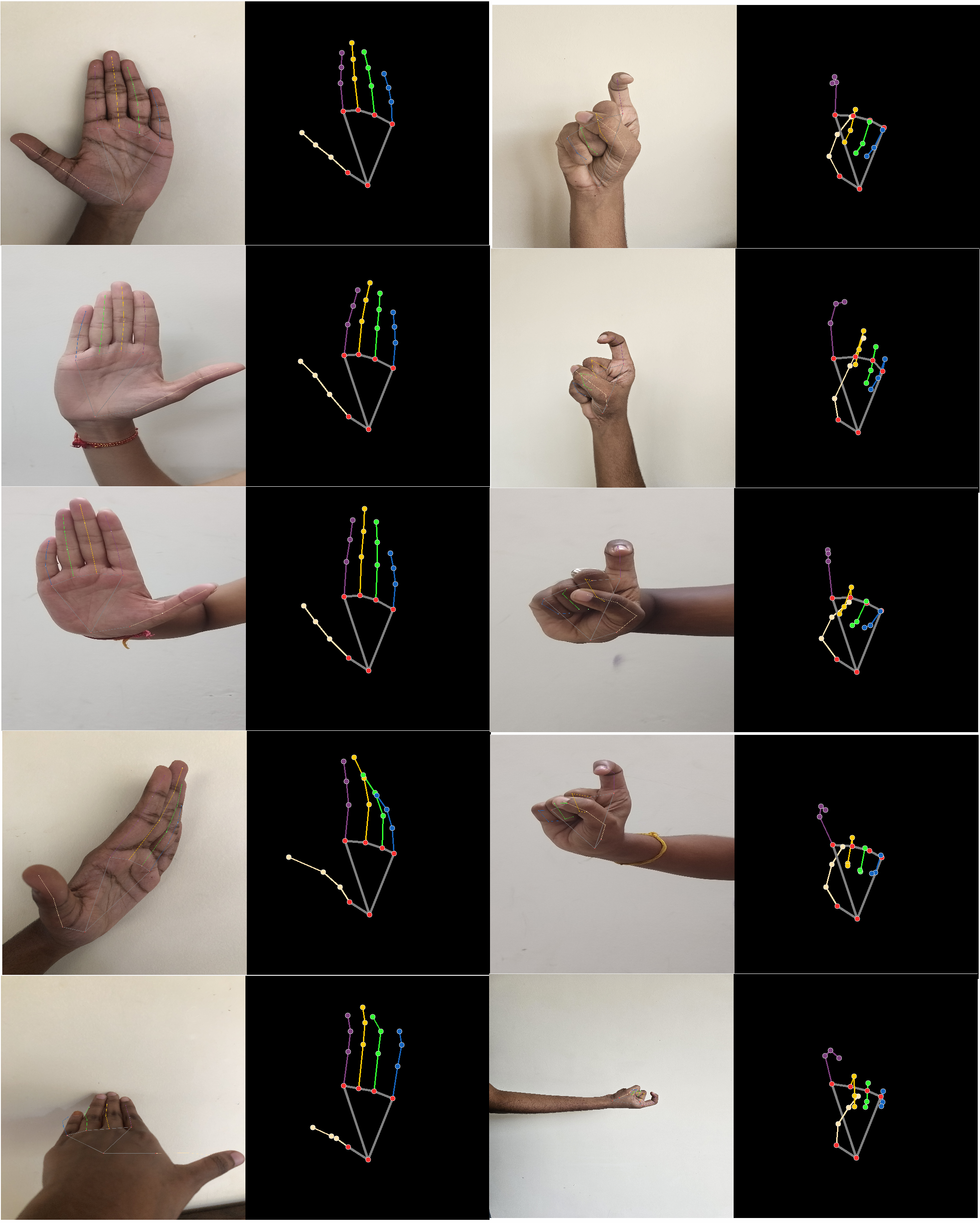}
    \caption{Normalization of hand landmarks: 2 Mudras shown with different hands(left and right), artists with different hand sizes, hands facing different directions, and varying zoom, all brought to a comparable representation.}
    \label{fig:norm}
\end{figure}

The 4 four points of the primary set can be fixed at the center of the frame, facing forward, and maintaining a fixed hand size. These points become part of the normalized coordinate system. The secondary set comprises the coordinate values acquired from pose estimation for these specific hand landmarks and belongs to the image coordinate system. By comparing the coordinates of landmarks in the image and normalized system, we determine the transformation needed to project the image landmarks onto the normalized coordinate system. The four points from the normalized coordinate system are represented as $P$(primary) and the corresponding four points from the image coordinate system are represented as $S$(secondary). 
\[
P = \begin{bmatrix}
    p_{0,x} & p_{0,y} & p_{0,z} \\
    p_{1,x} & p_{1,y} & p_{1,z} \\
    p_{2,x} & p_{2,y} & p_{2,z} \\
    p_{3,x} & p_{3,y} & p_{3,z} 
\end{bmatrix},       S = \begin{bmatrix}
    s_{0,x} & s_{0,y} & s_{0,z} \\
    s_{1,x} & s_{1,y} & s_{1,z} \\
    s_{2,x} & s_{2,y} & s_{2,z} \\
    s_{3,x} & s_{3,y} & s_{3,z} 
\end{bmatrix}
\]

These points refer to the following points of Fig \ref{fig:hand}:
\begin{itemize}
    \item $p_0$ and $s_0$: Wrist (point 0)
    \item $p_1$ and $s_1$: Thumb base (point 1)
    \item $p_2$ and $s_2$: Index base (point 5)
    \item $p_3$ and $s_3$: Pinky base (point 17)
\end{itemize}

We need to find a transformation matrix $T$ such that:

\[P = S \cdot T\]

The equation can be rearranged to solve for $T$ if the matrix $S$ is invertible:

\[T = P \cdot S^{-1}\]

In instances where the $S$ matrix derived from input values in the image coordinate system proves to be singular, rendering it non-invertible due to a determinant value of zero, recourse to linear algebra methodologies is usually recommended. Techniques such as least squares or singular value decomposition (SVD) offer viable solutions for determining the transformation matrix $T$. Alternatively, a simpler approach involves adjusting the input values slightly to induce a change in magnitude without distorting the hand shape, thereby ensuring a non-zero determinant. This adjustment could entail adding a small offset value to all x-coordinates (or similarly, to y or z coordinates).

\begin{algorithm}
\caption{Algorithm for Normalization of Coordinates obtained from pose estimation system}
\label{alg:normalization}
\KwIn{Hand Landmarks in the image coordinate system, $I \in \mathbb{R}^{21\times 3}$}
\KwOut{Vector representing hand Landmarks in the normalized coordinate system, $O \in \mathbb{R}^{1\times 63}$}
%\KwFunction{Normalize}{$I$}
\begin{enumerate}
    \item Initialize $P$

    \item Obtain the coordinates for the 4 landmarks in the image
\[
S = \begin{bmatrix}
    I_{0,0} & I_{0,1} & I_{0,2} \\
    I_{1,0} & I_{1,1} & I{1,2} \\
    I_{5,0} & I{5,1} & I{5,2} \\
    I_{17,0} & I_{17,1} & I_{17,2} 
\end{bmatrix}
\]
\item Compute transformation matrix $T=P\cdot S^{-1}$
\item $N = I \cdot T$
\item Flatten $N$, by concatenating the rows
\[ O = [N_{0,0}, N_{0,1}, N_{0,2}, N_{1,0}, N_{1,1}, N_{1,2}, {N_{3,0}, ..., N_{20,2}}]\]
\item \Return $O$
\end{enumerate}
\end{algorithm}

The values used to initialize $P$ in the implementation of Pose2Gest are as depicted in Fig \ref{fig:P-values} and the resultant matrix is given below:
\[
P = \begin{bmatrix}
    0.5 & 0.75 & 0 \\
    0.42 & 0.7 & -0.03 \\
    0.4 & 0.45 & -0.01 \\
    0.6 & 0.5 & -0.02 
\end{bmatrix}
\]
The x-value of wrist $P[0][0] = 0.5$ ensures horizontal placement of the hand at the middle of the frame and its y-value, $P[0][1] = 0.75$, places the wrist at the lower half of the image frame. Similarly, the other values also indicate the position being fixed for these four landmarks. As the distance between these points remains fixed, the normalization gives rise to a standard hand size and different input hand sizes would get scaled to match this required size. Moreover, the relative positions of these points depict a hand that is held with the thumb to the left. If the input image is left hand or hand facing another direction, the normalization rotates it to match the preset angle. Thus, the fixed values of $P$ determine the translation, scaling, and rotation of the input hands in the normalization process and can be set as per the requirements.

Fig \ref{fig:norm} depicts the input images and their normalized coordinate values for two different mudras in each column. It can be observed that the difference in the size of the hand, the angle of the hand, the zoom difference, the location of the hand in the image frame, etc get normalized effectively to be able to compare with each other and find a close match correctly.

Once the transformation matrix is thus obtained, it is applied to all points obtained from the model to project them into the normalized position, scale, and orientation. This normalization of coordinate features is performed on both training data(reference samples in the database) and incoming test data. As these are simple mathematical computations they can be incorporated without hampering performance in a real-time setting too.

\subsection{Vector Database}

Even though vector-based data processing is a field that has been studied for decades, we have seen increased popularity and the rise of dedicated database management systems over the last few years. Multiple production-ready vector database management systems are now available, capable of doing efficient similarity searches over large data collections\citep{pan2023survey}. 

In the proposed approach, under the most constrained setting of using only one training sample per class, it requires storing only 24 vectors in the database. As the size of the training set is increased to 5 images per class, it becomes 120, and for 10, it requires 240. Even then, the storage requirement is much less. However, robust database systems that are capable of indexing these vectors and scaling up easily to store more samples as employed as we would want to extend the system from mudras to more sophisticated and varied content like words of \textit{Kathakali} or other dance forms.  For the current implementation, the pgvector\citep{pgvectorpgvector_2024} database has been used which is an extension of Postgres DB for vector-based computing. Each normalized coordinate vector in the training data and its mudra label are stored in the database. 

The use of a database system is not mandatory for the implementation of the method. Alternatively using a K-Nearest Neighbour(KNN) implementation with n=1 (or appropriate values) or an Euclidean distance search implemented from scratch will also provide a similar result as long as the training dataset is small enough to be contained in memory. However, the use of a vector database ensures easier scaling up if the use case expands beyond tens or hundreds of classes.

\subsection{Classification}
When new test samples are obtained, either from videos or images, these are also normalized and checked against the data samples in the Database to find those with the closest similarity. The similarity score used is Euclidean distance as we are directly using the $x$,$y$, and $z$ coordinates in Euclidean space. A threshold is set to the similarity score to avoid very low matches. 

The system can be modified to include top $N$ ($N=3$, $5$, etc.) matches rather than just one match if the use case permits it. For instance, if the mudra prediction can be used as logits for a language model to predict the correct sequence used, considering more than one of the top predictions would be helpful to deduce the more probable sequence. When running on videos, A window of the last $10$ frames can be taken and the most frequent match among these $N*10$ values is taken as a valid prediction. This window-based approach allows fewer fluctuations in prediction results.

\begin{figure*}
    \centering
    \includegraphics[width=0.8\textwidth]{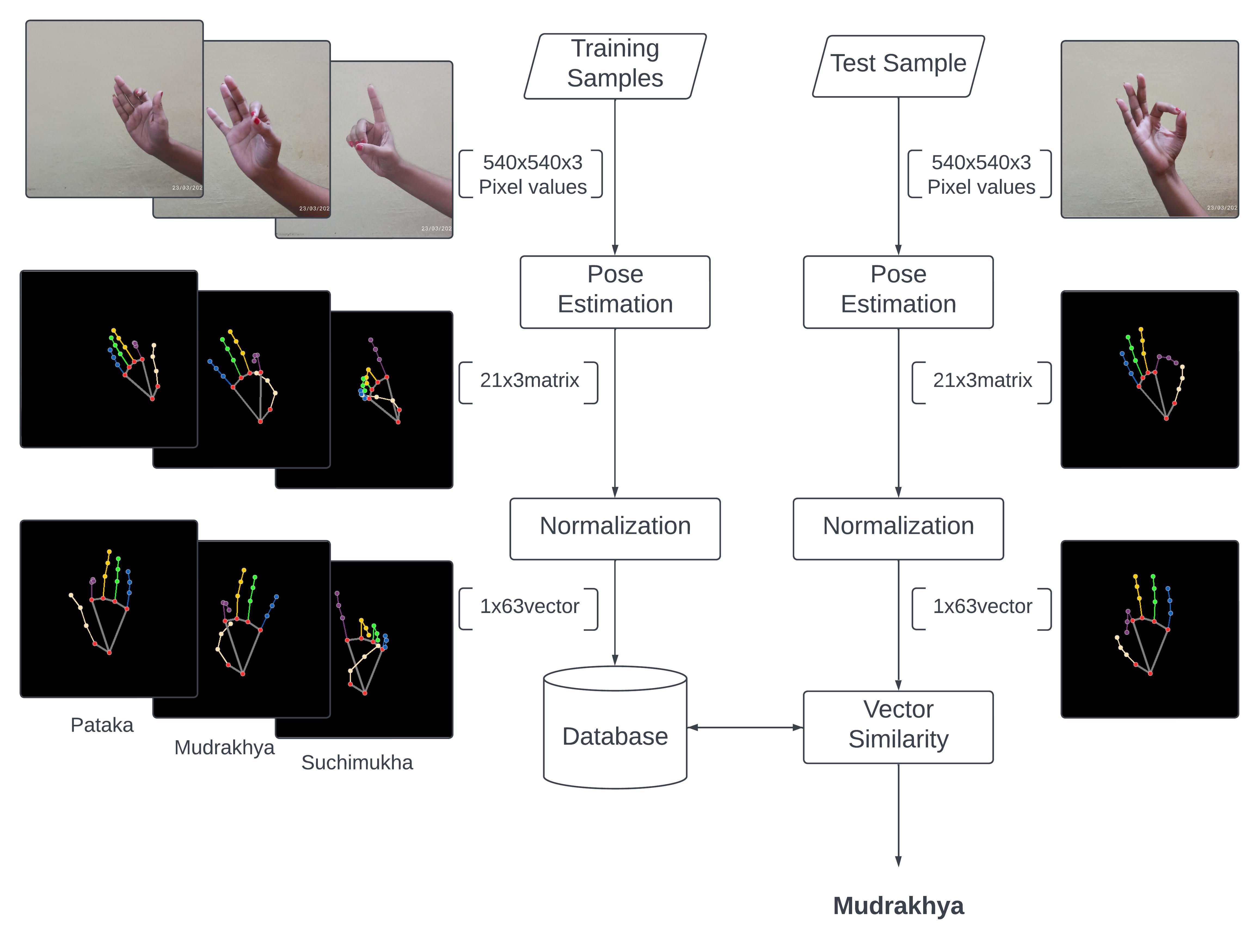}
    \caption{Dataflow in the Pose2Gest system: Training samples for $3$ classes are processed and saved in the database. The test sample is processed and compared against the database values to predict, the most similar match, \textit{"Mudrakhya"} as the class label. (Image size 540x540 given is only an example. The system does not require a fixed image size.)}
    \label{fig:dataflow}
\end{figure*}

Fig \ref{fig:dataflow} visualizes the Pose2Gest dataflow in a 3-class mudra recognition system with one sample per class in the training set. The single training samples representing each class get saved in the database along with class labels after pose estimation and normalization. The test sample, after similar data processing is compared against the references in the database, and the most similar class is determined. The differences in \textit{"Mudrakhya"} samples in the training set and test sample are minimized substantially after the processing and the resultant vectors are comparable even with a low dataset size of one sample per class.

\section{Experiments and Results}
We have conducted extensive experiments to compare the performance of our proposed method with that of existing works as well as to test its robustness upon throttling the data set size. The first set of experiments is conducted on publicly available datasets and we compare how our method performed against the reported accuracies from other works that use deep learning model training. The next two sets of experiments are done on our dataset by considering different data splits of varying training set sizes: 1 sample per class, 5 samples per class, 10 samples per class, and 80\% of the whole dataset.

The procedures followed in all the below-mentioned experiments are as per the steps discussed in the methodology. Pose estimation is done for all the training image samples, after which the coordinates values are normalized and stored in the database with corresponding class labels. Similarly, test set images are also subjected to pose estimation and normalization. After this, a vector similarity search based on Euclidean distance is run on the database and the topmost match is considered as the prediction. Even though there is no machine learning or neural network training involved in our approach we still refer to the reference data set as the training set in this article following the common practice in related literature.

\begin{table*}
\centering
\begin{tabular}{ lccc } 
 \hline
  Dataset & No. Of Classes & Existing Accuracy & Pose2Gest Accuracy\\
 \hline
  Kathakali Dataset I\cite{neelakanta_iyer-dataset} & 24 & 74\%\cite{bhavanam2020classification}, 79\%\cite{malavath2023classification} & 83\% \\ 
  Kathakali Dataset II\cite{saurav2021kathakalidataset} & 5 &- & 92\% \\ 
  Bharatanatyam\cite{parthasarathy2023novel} & 5 & 86.45\%\cite{parthasarathy2023novel} & 96\% \\ 
  Hasta Mudra Dataset & 24 & - & 92\% \\ 
 \hline
\end{tabular}
  \caption{Comparison of performance of the proposed method on existing public datasets. The scores reported by the original works are given as `Existing` and the scores observed using the proposed method, are given as `Pose2Gest'. The dataset split used is 80:20.}
  \label{tab:otherdatasets}

\end{table*}

\subsection{Tests on Existing Datasets}
The effectiveness of our proposed method is evaluated through comparisons with existing approaches on three datasets. The first two datasets pertain directly to the \textit{Kathakali} mudra recognition task. The first dataset, \citet{neelakanta_iyer-dataset}, is associated with prior work employing a neural network architecture for mudra recognition,\citet{bhavanam2020classification}. A second work, \citet{malavath2023classification} has also done CNN-based works on the same data. The proposed approach, which eliminates the need for model training and learning steps, is applied to this dataset to compare the effectiveness of the two methods directly. The second dataset, \citet{saurav2021kathakalidataset}, was also made available for the \textit{Kathakali} mudra Recognition task, but no published research has utilized it to date. Our work offers the first reported application of a mudra recognition method on this dataset. The third comparative analysis with \citet{parthasarathy2023novel}, involves a dataset containing Bharatanatyam mudras, a distinct Indian classical dance form. This experiment investigates the adaptability of our proposed method to similar tasks within a broader domain. By evaluating performance on this dataset, we assess the generalizability of our approach beyond the specific context of \textit{Kathakali} mudra recognition. The accuracies reported by existing works on these three datasets and that observed with the proposed Pose2Gest method are given in Table \ref{tab:otherdatasets}.

\subsubsection{Kathakali Dataset I}

The dataset published by \citet{neelakanta_iyer-dataset} along with their work(\citet{bhavanam2020classification}) contains images for 24 classes of \textit{Kathakali} mudras, which was used in CNN-based work. We have used the dataset and with a random 80:20 split, performed classification using our method. Upon examining the dataset it was noticed that the images are of 56x56 pixel size and hence of very low quality. The fingers are often indistinguishable and correct mudra recognition is difficult even manually for some samples. As a result of low-quality images pose estimation output is often incorrect in this dataset. All samples for which, pose estimation was able to detect a hand, even with errors, were considered for the experiment and obtained an accuracy of 83\% as reported in \ref{tab:otherdatasets}. This is a major improvement to the original work which reported an accuracy of only 74\% after using a CNN-based approach.

The separate attempt on the \textit{Kathakali} mudra classification(\citet{malavath2023classification}) using the same dataset(\citet{neelakanta_iyer-dataset}) employed a naive Bayes classification technique and CNN model and reported a better performance using CNN, which is 79\%. Our result of 83\% has surpassed this accuracy as well.

\subsubsection{Kathakali Dataset II}

A 5-class dataset is available on Kaggle\citep{saurav2021kathakalidataset} for \textit{Kathakali} mudras. It contains 5 mudras, namely \textit{`mudrakhya', `pataka', `kataka', `kartari mukha', and  `musti',} which is a small subset of the complete set of 24 mudras. This dataset contains images with higher pixel clarity and more hand orientations. In certain image samples, due to the shape of the mudra and the hand orientation, not all fingers are visible and hence mudra is not clear, though. This dataset was also used and with an 80:20 split on the data for the training and the test sets, we applied our method for mudra classification. Here an increased accuracy of 92\% is observed, which can be attributed to the better image quality and reduced ambiguity due to fewer classes. There are no other published works that report accuracy on this dataset.

\subsubsection{Bharatanatyam Dataset}
\textit{Bharatanatyam} is another classical dance popular in India, which also has the concept of mudras or hand gestures. But it uses a slightly different set of mudras, with hand shapes and names differing from the \textit{Kathakali} mudras. In the works of \citet{parthasarathy2023novel}, a benchmark dataset was published for \textit{Bharatanatyam} mudras, which includes short video clips for 5 mudra classes. Image frames are obtained from these video clips at the rate of FPS=5. From the available videos in the published dataset, we obtained 916 images for the experiment, even though the original work has reported using 6000 images. Even on a reduced data set size, our approach surpasses the reported accuracy of 86\%, by exhibiting a 96\% accuracy on this 5-class classification problem.

\begin{table*}
\centering
    \begin{tabular}{lclc}
    \hline
    Mudra Class & No. of Samples & Mudra Class & No. of Samples \\
    \hline
	Pataka & 38 & Mukura & 40 \\
	Mudrakhya & 39 & Bhramara & 36 \\
	Kataka & 37 & Suchimukha & 37 \\
	Mushti & 38 & Pallava & 36 \\
	Kartarimukha & 42 & Tripataka & 40 \\
	Sukatunda & 39 & Mrigasirsha & 39 \\
	Kapittha & 38 & Sarpasiras & 39 \\
	Hamsapaksha & 40 & Vardhamanaka & 38 \\
	Sikhara & 38 & Arala & 39 \\
	Hamsasya & 43 & Urnanabha & 33 \\
	Anjali & 41 & Mukula & 38 \\
	Ardhachandra & 42 & Katakamukha & 38 \\
    \hline
    \end{tabular}
    \centering
    \caption{The 24 classes and the number of samples per class in the Hasta Mudra dataset. These mudras are as per \textit{Hastalakshana Deepika}\cite{hastalakshana}.}
    \label{tab:Hastamudra}
\end{table*}
\subsection{Tests on Hasta Mudra Dataset}

As part of this work, we have developed a 24-class dataset for \textit{Kathakali} mudras involving 8 participants, different angles of hand, left and right hands, different zoom, lighting, and such capturing conditions. 
The dataset is publicly available at \url{https://github.com/kavitharaju/KathakaliMudraDataset}. The classes included and the number of samples in each class are given in Table \ref{tab:Hastamudra}. All the data splits used for the following experiments are made available in the dataset repository.

To observe the performance of the approach in a data-constrained situation we have throttled our training set size to extremely small sizes and tested it against the remaining samples. The experiment procedures followed in this set of experiments are also the same as the previous ones, which involved pose estimation, normalization, and vector similarity-based prediction. 

The experiments used training sets of the following sizes:
\begin{enumerate}
    \item One sample per class
    \item Five samples per class
    \item 10 samples per class
    \item 80\% of the full dataset
\end{enumerate}

\begin{table}
  \centering
  \begin{tabular}{lcc}
    \hline
    %\multirow{Training Dataset} & \multicolumn{2}{c}{All 24 classes} & \multicolumn{2}{c}{Only 10 classes} \\
    %\cline{2-5}
     %       & Exp No. & Accuracy & Exp No. & Accuracy \\
     Training Set & On 24 classes & On 10 classes \\
    \hline
    1 sample  & 63\% & 81\%\\
    5 samples  & 75\% & 91\% \\
    10 samples  & 83\% & 91\% \\
    80\% of dataset & 92\% & 95\%\\
    \hline
  \end{tabular}
  \caption{Experiment results on Hasta Mudra dataset with varying training set sizes. Scores are given for the full 24 class dataset as well as the selected 10 unambiguous class datasets.}
  \label{tab:ourdatasetsplits}

\end{table}

The accuracies observed for each of these training set sizes are reported in Table \ref{tab:ourdatasetsplits}, which shows reasonable performance even at such low dataset sizes. The data split for some of the experiments has a test set size which is several folds of the training set size which is not common for Machine learning or deep learning evaluations. It is because of the tight dataset constraint we wanted to enforce, that such tests were conducted. Table \ref{tab:datasize} lists the sizes of training sets and test sets in each of the experiments.

\begin{table}
\centering
\begin{tabular}{ lcc } 
 \hline
 Experiment & Train set  & Test set \\
 Description & Size & Size \\
 \hline
 Kathakali Dataset I\cite{neelakanta_iyer-dataset} & 485 & 169 \\ 
 Kathakali Dataset II\cite{saurav2021kathakalidataset} & 991 & 245 \\ 
 Bharatanatyam Dataset\cite{parthasarathy2023novel} & 732 & 184 \\ 
 Hasta Mudra Dataset &  &  \\ 
 \hspace{1cm}24-class 1 sample & 24 & 904 \\
 \hspace{1cm}24-class 5 samples & 120 & 808 \\ 
 \hspace{1cm}24-class 10 samples & 240 & 688 \\ 
 \hspace{1cm}24-class 80:20 split & 742 & 186 \\ 
 \hspace{1cm}10-class 1 sample & 10 & 381 \\ 
 \hspace{1cm}10-class 5 samples & 50 & 341 \\ 
 \hspace{1cm}10-class 10 samples & 100 & 291 \\ 
 \hspace{1cm}10-class 80:20 split & 315 & 76 \\ 

 \hline
\end{tabular}
  \caption{Sizes of training and test splits in the experiments}
  \label{tab:datasize}
\end{table}

\subsection{Tests on unambiguous classes}

The hand shapes of certain pairs of mudras can be very similar to each other like in the case of \textbf{Tripataka} and \textit{Anjlai} or \textit{Mushti} and \textit{Katakamukha}. This can cause a certain amount of ambiguity in the task of mudra recognition and can have a major impact on the samples where hand orientation makes their difference minimal. Several works \citep{parthasarathy2023novel} \citep{saurav2021kathakalidataset} explore the performance of their methods on a simplified problem including only a subset of mudras that have substantial differences in the hand shape. To explore this performance difference we have reduced our dataset, the hasta mudra, and task to a 10-class problem including only the following 10 mudras: \textit{`Pataka', `Mudrakhya', `Sukatunda', `Hamsapaksha', `Sikhara', `Ardhachandra', `Mukura', `Arala', `Mukula', Katakamukha'}.

The training set sizes used in these experiments were also similar to the previous set of experiments: 
\begin{enumerate}
    \item One sample per class
    \item Five samples per class
    \item 10 samples per class
    \item 80\% of the full dataset
\end{enumerate}

Table \ref{tab:ourdatasetsplits} shows the accuracies obtained in these experiments as the score on 10 classes. Choosing this subset has reduced the complexity of the problem giving higher accuracy, especially in few-shot experiments. These scores are better comparable with the scores of \textit{Kathakali} Dataset II and \textit{Bharatanatyam} in Table \ref{tab:otherdatasets}, as these datasets only contain 5 classes each.

\section{Implementation}

The proposed methodology has been implemented as a web-based software tool and is available at \url{https://mudraresearch.iiitkottayam.ac.in/}. The interface allows users to upload image files or capture pictures using a webcam that contains images of people showing any of the \textit{Kathakali} mudras with their hand. These images are processed and identified mudras will be displayed to the user. The implementation uses Python at the back end and React at the front end. The tool is also backed by the data of Hasta Mudra Dataset.

The tool serves as a showcase for the efficiency of the approach and bridges research progress with artistic communities. This facilitates improved collaboration, offering a clearer comprehension of the research's potential benefits to stakeholders and furnishing researchers with prompt, firsthand feedback on the system's performance across various real-world test scenarios.

Crowdsourcing data collection and annotation is another purpose for the tool. Every trial on the app also asks the user to validate the prediction, stating whether the automated mudra prediction was correct or wrong. This feedback is of high value to us:
\begin{itemize}
    \item It is a human evaluation of the system.
    \item The images provided along with the correct labels marked, become human-annotated data for improving the data set size and thus system performance. As the system is designed to work on low-resource domains, a pilot implementation can be provided with a few samples per class, and user trials and feedback can be used for building up the dataset.
\end{itemize}

The tool has also integrated the Pose2Gest method into addressing the Sign Language Recognition challenge. Gesture recognition plays a critical role in sign language processing, a field characterized by intricate vocabularies and gestures far more elaborate than those found in traditional art forms. Within sign language, fingerspelling represents a subset of vocabulary, encompassing hand signs for the 26 English letters and utilized for spelling names, places, new terms, etc. This specific issue is framed as a 26-class classification task and integrated into the tool using a similar approach applied to mudras, with adaptations for bilateral hand gestures as certain signs in Indian Sign Language involve both hands.

\section{Discussion}
The proposed approach can attain the performance of deep learning models trained or finetuned for specific tasks, which is evident in the experiment results. Furthermore, it can give satisfactory results even with very small data available for training a baseline system in a new task, which is critical when the target domains face data scarcity problems and the threat of endangerment. Though the work has focused on \textit{Kathakali}, the proposed method can be applied to other similar tasks easily as demonstrated by the \textit{Bharatanatyam} mudra recognition experiments and Sign language finger spelling recognition implemented in the tool. Moreover, the dataset used is directly applicable to the artforms \textit{Kutiyattam, Mohiniyattam,}  and \textit{Krishnanattam}, as the mudras used in these are all based on \textit{Hasta Lakshana Deepika}\cite{hastalakshana}.

The experiments were done on isolated, cropped images to be able to compare effectively with other works. The pre-trained pose detection model we use is capable of performing good-quality pose estimation on images with full body and background details, continuous videos, real-time streams, and even on mobile devices\citet{lugaresi2019mediapipe}. As the additional computation of normalization and vector similarity done on top of it are of negligible computational costs, the method easily extends to a continuous, real-time, and cross-platform solution. However, the system has not been tested extensively on images or videos of \textit{Kathakali} in costume. This could posses additional challenges in accurate pose estimation due to unusual attire and face paintings used in \textit{Kathakali} which the general purpose model might not have encountered in its training data. In such scenarios, works like the Kathakali Framework\cite{bulani2022framework} which developed a specialized pose estimation system could be employed.

The application of \textit{Kathakali} mudra detection comes as a first step towards \textit{Kathakali} interpretation. \textit{Kathakali}'s interpretation has its relevance as the performance includes signed dialogues that are not intelligible for a novice in the audience. But applications of the mudra recognition system come in other forms also. It can serve as a pedagogical tool applying the system to detect select mudras and recommend ideal orientations. Another application is in tagging unannotated data for creating larger corpora as our target domain has the problem of data scarcity.

\section{Conclusion}
The main contribution of this work is the novel approach for image classification tasks involving hand gestures such as dance mudra recognition in Indian classical dance forms which can give satisfactory results even with a very constrained dataset size. This is critical as there are several similar applications for this in art forms and Sign languages where the foremost constraint for deep learning approaches is their huge data requirements. Our method could help with a jump start in such tasks which can later be employed for creating more data from unannotated corpora. The comparative study of the proposed methodology with existing works, the Hasta Mudra dataset, and the web tool that implements the proposed approach are the additional contributions of this work.

Extending the system for \textit{Kathakali} word recognition is one of the future directions planned for the work. A \textit{Kathakali} word is formed using one or more of the mudras held or moved in specified postures. These span over multiple frames. The current mudra recognition is limited to single frame level classification, but the bigger task requires us to work with videos involving more context. This calls for more challenges and exploration of solutions. We plan to employ full-body pose estimation in place of only hands and figure out effective ways to segment videos based on pose changes. Again the aim is to work with as minimum data as possible.

Other future directions identified are: Applying the same solutions to different problems like Koodiyattam and other dance forms. Applying similar techniques in word-level sign language recognition as in \textit{Kathakali}, to be able to adapt the solutions easily to different sign languages and their variants with minimum data dependency. 

\section{Acknowledgments}
We extend our heartfelt gratitude to the volunteers who provided guidance and contributed to the creation of the Hasta Mudra dataset, which is a key component of this project's publication. Among the contributors are artists from Kerala Kalamandalam, members of the cultural club at the Indian Institute of Information Technology-Kottayam, and family members. Additionally, we wish to express our appreciation to the interns and Bridge Connectivity Solutions Pvt. Ltd. for their valuable contributions and support in the development of the web application.

% \section*{References}
% \nocite{*}
\bibliographystyle{alpha}
\bibliography{KathakaliMudras}
% \appendix

% \section{Example Appendix}
% \label{sec:appendix}

% This is an appendix.

\end{document}